\documentclass{article}
\usepackage{iclr2026_conference,times}

\usepackage{graphicx}
\usepackage{amsmath,amssymb}
\usepackage{booktabs}
\usepackage{array}
\usepackage{multirow}
\usepackage{enumitem}
\usepackage{nicefrac}
\usepackage{upgreek}

\usepackage{algorithm}
\usepackage{algorithmic}

\usepackage{amsmath,amsfonts,bm}

\def\eqref#1{equation~\ref{#1}}
\def\1{\bm{1}}

\DeclareMathAlphabet{\mathsfit}{\encodingdefault}{\sfdefault}{m}{sl}
\SetMathAlphabet{\mathsfit}{bold}{\encodingdefault}{\sfdefault}{bx}{n}

\def\gD{{\mathcal{D}}}

\usepackage[bookmarks=true]{hyperref}
\usepackage{url}

\usepackage{tikz}
\usetikzlibrary{arrows.meta, positioning, calc}
\usepackage{amsmath}
\usepackage{graphicx}

\title{Bridging the Gap: Enabling Soft Actor Critic\\for High Performance Legged Locomotion\thanks{This paper is a technical report accompanying the open-source release of RSL-RL-SAC, an extension of the widely used RSL-RL reinforcement learning library.}}

\author{%
  Gianluca Sabatini, \quad Chenhao Li, \quad Marco Hutter \\
  ETH Zurich, Switzerland \\
  \texttt{\{gsabatini, chenhli, mahutter\}@ethz.ch}
}

\iclrfinalcopy

\begin{document}

\maketitle
\lhead{}

\begin{center}
\vspace{-2em}
\textcolor{blue}{\url{https://sabagian.github.io/sac_release_project/}}
\end{center}

\begin{abstract}
Proximal Policy Optimization (PPO) has become the de facto standard for training legged robots, thanks to its robustness and scalability in massively parallel simulation environments like IsaacLab. However, its on-policy nature makes it inherently sample-inefficient, preventing its use for continuous adaptation and fine-tuning on real hardware. Soft Actor-Critic (SAC), by contrast, is an off-policy algorithm that can reuse past experience, making it a natural candidate for sim-to-real transfer workflows where the same algorithm can be used both in simulation and for online learning on the real robot. Despite these advantages, SAC has consistently failed to match PPO's empirical performance in massively parallel training settings. This work identifies the root causes of this gap and introduces targeted modifications, covering policy initialization, timeout-aware critic targets, and multi-step return estimation, that enable SAC to train stably at scale. Evaluated across multiple legged robot platforms and diverse locomotion tasks, our approach closes the performance gap with PPO entirely.

\end{abstract}

\begin{figure*}[t]
  \centering
  \includegraphics[width=\linewidth]{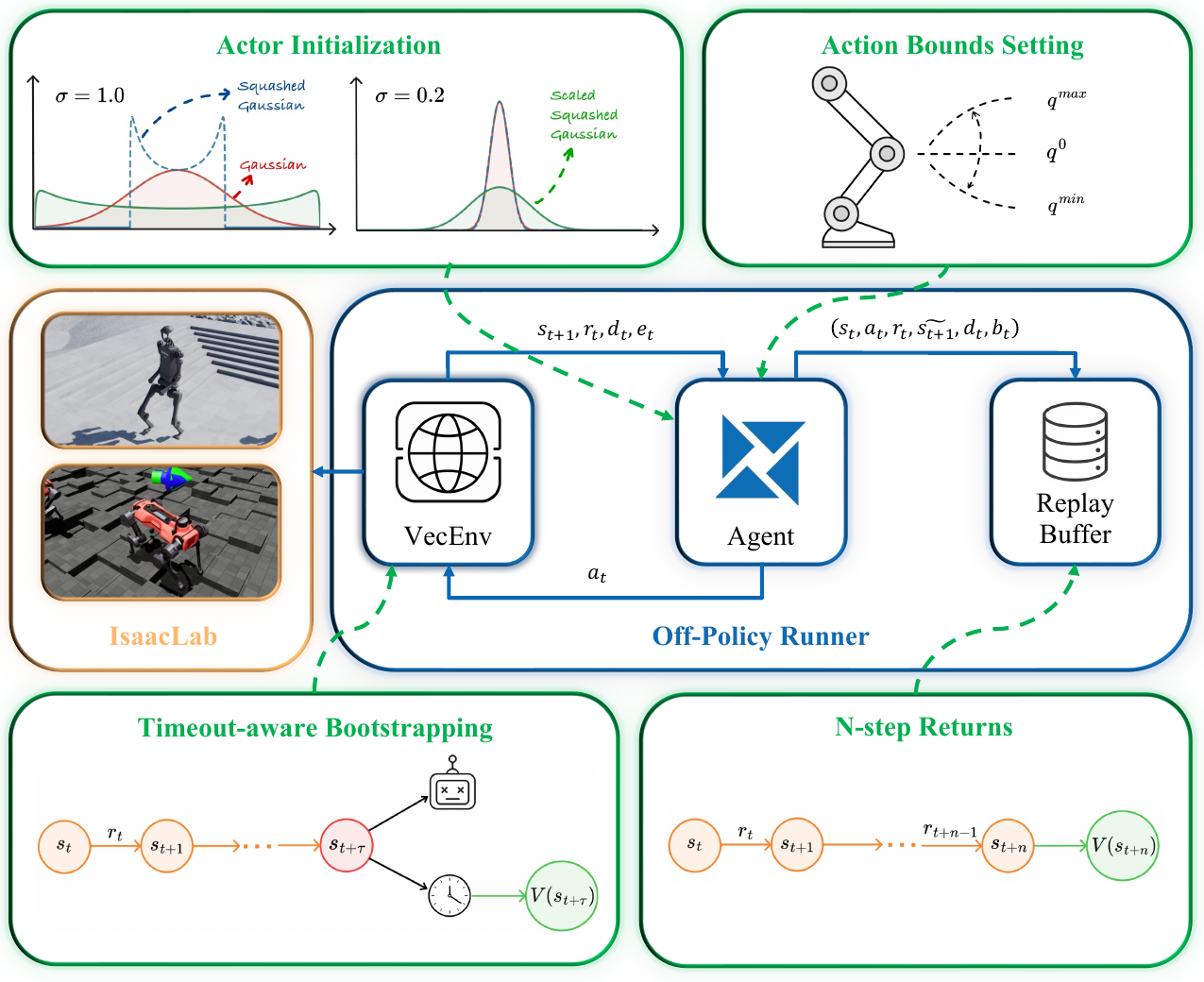}
  \caption{Overview of our RSL-RL-SAC framework for training legged locomotion policies in IsaacLab.
  Four modifications are highlighted: actor initialization with
  small $\sigma$ to speed up training, action bounds from robot joint limits,
  timeout-aware bootstrapping that distinguishes timeouts from failures, and $n$-step
  returns for faster reward propagation.}
  \label{fig:main_fig}
\end{figure*}

\section{Introduction}
\label{sec:introduction}

Legged robots operating in unstructured real-world environments must continuously adapt 
to novel terrain and perturbations, making online learning on hardware a long-term 
objective of the field \citep{smith2021leggedrobotslearningfinetuning, 
as2026matterssimulationonlinereinforcement}. Massively parallel simulation frameworks 
\citep{makoviychuk2021isaac, zakka2025mujoco, nvidia2025isaaclabgpuacceleratedsimulation} 
have dramatically reduced the time needed to train locomotion policies, enabling 
remarkable results across a wide range of platforms and tasks 
\citep{rudin2022learning, agarwal2023legged, cheng2024extreme, zhuang2024humanoid}. 
Within this paradigm, Proximal Policy Optimization (PPO) 
\citep{schulman2017proximal} has become the dominant training algorithm, owing to 
its stability and scalability across thousands of parallel environments.

Despite these successes, PPO's on-policy nature poses a fundamental limitation: 
it can only learn from data collected by the current policy, discarding all 
transitions after each update. This makes PPO inherently sample-inefficient, a property that is tolerable in simulation, where data is cheap, but becomes 
a critical bottleneck for real-world deployment. On physical hardware, where 
rollouts are slow, costly, and constrained by real-time execution, sample 
efficiency is essential. Robots that must keep learning after deployment cannot 
afford to discard their experience \citep{as2026matterssimulationonlinereinforcement, smith2021leggedrobotslearningfinetuning, li2025roboticworldmodelneural}.

Off-policy algorithms offer a principled solution to this problem. By storing 
transitions in a replay buffer and reusing them across many gradient updates, 
they can achieve substantially higher sample efficiency than on-policy methods 
\citep{as2026matterssimulationonlinereinforcement}. In particular, Soft 
Actor-Critic (SAC) \citep{haarnoja2018soft2}, a maximum entropy off-policy 
algorithm, is a natural candidate for bridging simulation training and real-world 
fine-tuning: a single algorithm could be used both to pretrain in simulation and 
to adapt on hardware, without requiring the massive parallelism that PPO depends on 
\citep{smith2021leggedrobotslearningfinetuning}. Yet despite these compelling 
properties, SAC and other off-policy methods have consistently failed to match 
PPO's empirical performance in massively parallel training environments \citep{li2026uncertaintyawareroboticworldmodel}.

The reasons for this gap have been recently investigated 
\citep{raffin2025isaacsim, shukla2025fastsac, seo2025learningsimtorealhumanoidlocomotion}, 
revealing that the massively parallel simulation setting exposes fundamental 
incompatibilities between standard SAC and this training regime. These include 
miscalibrated action space exploration arising from unbounded environment action 
spaces, incorrect handling of episode truncations that biases critic targets, 
and reward propagation that is too slow for stable learning on rough terrain. 
Existing solutions either require task-specific reward engineering 
\citep{seo2025learningsimtorealhumanoidlocomotion}, rely on distributional 
critics that introduce additional complexity and variance 
\citep{seo2025fasttd3}, or demand extensive hyperparameter tuning that 
limits generalization \citep{raffin2025isaacsim, shukla2025fastsac}. 
None of these works systematically closes the gap with PPO across a broad 
range of tasks and robot platforms using unmodified reward functions.

In this work, we identify and address the core sources of instability that 
prevent SAC from scaling in massively parallel simulation, and propose a set 
of principled modifications that require no task-specific tuning. Our approach 
is evaluated against the RSL-RL PPO implementation \citep{schwarke2025rsl} 
across seven legged robot platforms, from quadrupeds to humanoids, on 
velocity tracking tasks over rough terrain, using the same reward functions 
originally designed for PPO.

Our main contributions are:
\begin{itemize}
    \item We identify the key failure modes of SAC in massively parallel 
    simulation: miscalibrated initial exploration due to action space mismatch, 
    biased critic targets from incorrect truncation handling, and slow reward 
    propagation on rough terrain.
    \item We demonstrate that the resulting algorithm closes the performance 
    gap with PPO across all evaluated tasks and platforms, using a single 
    hyperparameter configuration and unmodified PPO reward functions.
    \item We release our implementation built on top of RSL-RL, extended 
    with Random Network Distillation, symmetry-based data augmentation, 
    and multi-GPU training support.
\end{itemize}

\section{Related Work}
\label{sec:related_work}

\paragraph{Massively parallel simulation for legged locomotion.}
GPU-accelerated simulation frameworks have become the standard substrate for 
training RL-based locomotion controllers 
\citep{makoviychuk2021isaac, mittal2023orbit, zakka2025mujoco, 
nvidia2025isaaclabgpuacceleratedsimulation}. By running thousands of 
environments in parallel on a single workstation, these frameworks have enabled 
remarkable results across a wide range of robot morphologies and tasks, including 
quadrupedal locomotion over challenging terrain, dexterous manipulation, and 
humanoid parkour 
\citep{rudin2022learning, agarwal2023legged, cheng2024extreme, 
singh2024dextrah, zhuang2024humanoid}. However, the dominant training algorithm 
across all of these works is PPO, an on-policy method whose sample inefficiency 
is tolerable in simulation but becomes a fundamental limitation for online 
learning on physical hardware.

\paragraph{Sim-to-real transfer and online fine-tuning.}
Policies trained purely in simulation often fail to generalize to the full 
diversity of real-world conditions, motivating approaches that combine 
simulation pretraining with online adaptation on hardware. 
\citet{smith2021leggedrobotslearningfinetuning} demonstrated that off-policy 
RL can enable continuous performance improvement when transitioning from 
simulation to real environments, showing that a replay buffer populated in 
simulation can bootstrap efficient learning on the physical robot. In a related 
line of work, \citep{smith2023walk} showed that off-policy RL can learn 
locomotion policies directly on hardware within minutes, without requiring 
massive environment parallelism. More 
recently, \citet{as2026matterssimulationonlinereinforcement} systematically 
studied what factors matter most for this sim-to-real fine-tuning paradigm, 
highlighting sample efficiency as a first-order concern. These works motivate 
the use of off-policy algorithms such as SAC, which can reuse past experience 
and operate effectively even without massive environment parallelism.

\paragraph{Scaling off-policy RL in massively parallel simulation.}
Despite the appeal of off-policy methods for robotics, scaling them to 
massively parallel simulation settings has proven surprisingly difficult. 
Early work by \citet{li2023parallel} and \citet{gallici2024simplifying} 
showed that off-policy algorithms can be adapted to this regime, but did not 
match PPO's performance on locomotion tasks. More recently, 
\citet{raffin2025isaacsim} identified that 
a key source of failure is the mismatch between SAC's squashed Gaussian 
action distribution and the overly large action space bounds commonly used 
in IsaacLab environments and, as \citet{shukla2025fastsac}, showed that SAC can be made to work through 
careful hyperparameter tuning. In a parallel line of work, 
\citet{seo2025learningsimtorealhumanoidlocomotion} and 
\citet{seo2025fasttd3} introduced FastSAC and FastTD3, achieving strong 
results on humanoid locomotion tasks. However, these methods rely on 
task-specific reward engineering, limiting their generalization to new 
platforms, and employ distributional critics \citep{bellemare2017distributional} 
that introduce additional complexity in critic training and can increase 
variance in actor updates. Furthermore, they typically involve a large number 
of sensitive hyperparameters, requiring extensive tuning to achieve good 
performance.

In contrast to prior works that achieve competitive performance through 
extensive hyperparameter tuning or reward engineering, our 
approach provides a systematic analysis of the root causes of SAC's 
failure in massively parallel simulation and proposes principled 
modifications that address each failure mode directly. Built on top of 
RSL-RL, a widely adopted library for robot learning, our implementation 
is evaluated across a broad range of legged platforms spanning both 
quadrupeds and humanoids.

\section{Method}
\label{sec:method}

The modifications introduced in this work address three distinct failure modes 
of standard SAC in massively parallel simulation: miscalibrated initial 
exploration arising from unbounded action spaces, biased critic targets caused 
by incorrect handling of episode truncations, and slow reward propagation on 
rough terrain. We present each component in turn, preceded by a description 
of the SAC objective and policy parameterization used throughout.

\subsection{Preliminaries}
\label{sec:preliminaries}

We model the environment as a Markov Decision Process (MDP) defined by the 
tuple $\mathcal{M} = (\mathcal{S}, \mathcal{A}, p, r)$, where 
$\mathcal{S} \in \mathbb{R}^s$ and $\mathcal{A} \in \mathbb{R}^d$ represent 
the state and action spaces, and $p$ and $r$ denote the unknown transition 
and reward functions, respectively. In the maximum entropy RL framework, the 
agent seeks a policy $\pi^*:\mathcal{S}\rightarrow\mathcal{A}$ that maximizes 
the expected return augmented by a weighted entropy bonus:
\begin{align}
\pi^*
=
\arg\max_{\pi}
\sum_{t}
\mathbb{E}_{(s_t,a_t)\sim \rho_\pi}
\left[
r(s_t,a_t)
+
\alpha \mathcal{H}\!\left(\pi(\cdot \mid s_t)\right)
\right].
\end{align}
In the infinite-horizon setting, a discount factor $\gamma \in (0, 1)$ 
ensures the objective remains finite. We use $N_e$ to denote the number 
of parallel environments and $N_s$ the number of steps collected per 
environment per iteration.

The critic is trained by minimizing the soft Bellman residual over an 
ensemble of $N$ Q-functions, using target networks updated via exponential 
moving average and clipped double Q-learning to mitigate overestimation 
bias~\citep{fujimoto2018addressing}:
\begin{equation}
\begin{aligned}
J_Q(\{\theta_i\}_{i=1}^N)
&=
\mathbb{E}_{(s_t,a_t)\sim\mathcal{D}}
\left[
\frac{1}{N}
\sum_{i=1}^N
\left(
Q_{\theta_i}(s_t,a_t)
-
Q^{\text{target}}_{\bar{\theta}}
\right)^2
\right],\\[0.5em]
Q^{\text{target}}_{\bar{\theta}}
&=
r(s_t,a_t)
+
\gamma \mathbb{E}_{s_{t+1}\sim p}
\big[
V_{\bar{\theta}}(s_{t+1})
\big],\\[0.5em]
V_{\bar{\theta}}(s_{t+1})
&=
\mathbb{E}_{a_{t+1}\sim\pi}
\left[
\min_{i\in\{1,\dots,N\}} Q_{\bar{\theta}_i}(s_{t+1},a_{t+1})
-\alpha \log \pi(a_{t+1}\mid s_{t+1})
\right].
\end{aligned}
\end{equation}
The actor is trained to maximize the expected return plus entropy using 
the reparameterization trick:
\begin{align}
J_\pi(\phi)
=
\mathbb{E}_{s_t\sim\mathcal{D},\, \epsilon_t\sim\mathcal{N}}
\left[
\alpha \log \pi_\phi
\big(
f_\phi(\epsilon_t; s_t)
\mid s_t
\big)
-
\min_{i\in\{1,\dots,N\}} Q_{\theta_i}
\big(
s_t, f_\phi(\epsilon_t; s_t)
\big)
\right].
\end{align}
The temperature $\alpha$ is tuned automatically by optimizing:
\begin{align}
J(\log \alpha)
=
\mathbb{E}_{a_t\sim\pi_t}
\left[
- (\log \alpha)
\left(
\log \pi_t(a_t\mid s_t)
+
\bar{\mathcal H}
\right)
\right],
\end{align}
where $\bar{\mathcal{H}}$ is the target entropy. We found automatic 
temperature tuning to be beneficial for both training stability and final 
performance.

\subsection{Policy Parameterization and Action Bounds}
\label{sec:policy_param}

The policy is parameterized as a state-dependent Gaussian. The actor network 
outputs a mean $\mu_\phi(s) \in \mathbb{R}^d$ and a raw log-standard-deviation 
$\tilde{\ell}_\phi(s) \in \mathbb{R}^d$, which is clamped and exponentiated to 
give $\sigma_\phi(s) = \exp(\mathrm{clip}(\tilde{\ell}_\phi(s), \ell_{\min}, 
\ell_{\max}))$. The latent action is sampled via the reparameterization trick,
$x = \mu_\phi(s) + \sigma_\phi(s) \odot \varepsilon$, $\varepsilon \sim 
\mathcal{N}(0, I)$, and passed through a $\tanh$ squashing 
function~\citep{haarnoja2018soft2} to produce $u \in (-1,1)^d$.

A critical issue in applying SAC to IsaacLab environments is the mismatch 
between the environment's action space definition and the effective range 
of motion required by the task. Depending on the environment wrapper, 
action bounds may be left unscaled, constraining policy outputs to 
$(-1, 1)^d$ and producing insufficiently small joint displacements, or 
set to arbitrarily large values that far exceed the range used by a 
trained policy. In the latter case, SAC's squashed Gaussian rescales its 
outputs to fill the entire declared action space, spreading probability 
mass near the action limits and inducing erratic exploratory behavior from 
the first iteration~\citep{raffin2025isaacsim}. In either case, the 
mismatch between the declared action space and the task-relevant range of 
motion causes the initial exploration distribution to be poorly calibrated, 
impeding learning from the outset. We address this by 
deriving tight action bounds directly from the robot's joint configuration, following a similar physics-based approach to ~\citep{seo2025learningsimtorealhumanoidlocomotion}, 
without requiring a pretrained policy. Specifically, the environment 
processes policy outputs as $a_t^p = s \odot a_t + b^e$, where $b^e$ is 
the default joint configuration and $s$ is the task-specific action scale. 
For each joint $j$, we compute the distance from the default position to 
the \textit{soft} joint limits, $r_j^{\pm} = |q_j^{\min/\max} - q_j^0|$, 
and rescale by the action-manager scale to obtain per-joint policy output 
bounds $a_j^{\min} = -r_j^-/s$, $a_j^{\max} = r_j^+/s$. Using soft rather 
than hard limits is important: hard limits introduce suboptimal regions 
that degrade performance. Defining $b = \frac{1}{2}(a_{\max} + a_{\min})$ 
and $c = \frac{1}{2}(a_{\max} - a_{\min})$, the final action and 
log-probability are:
\begin{align}
f_\phi(\varepsilon; s)
&=
b + c \odot \tanh\!\left(\mu_\phi(s) + \sigma_\phi(s) \odot \varepsilon\right),
\\
\log \pi_\phi(a \mid s)
&=
\sum_{j=1}^{d}
\log \mathcal{N}\!\left(x_j; \mu_{\phi,j}, \sigma_{\phi,j}^2\right)
-
\sum_{j=1}^{d}
\log\!\left(1 - \tanh^2(x_j)\right)
-
\sum_{j=1}^{d}
\log c_j,
\end{align}
where the log probability correction follows from the change of variables 
formula applied to the element-wise $\tanh$ transformation and scaling.

\subsection{Actor Initialization}
\label{sec:actor_init}

\begin{figure}[t]
\centering
\begin{minipage}{0.45\textwidth}
\centering
\includegraphics[width=\linewidth]{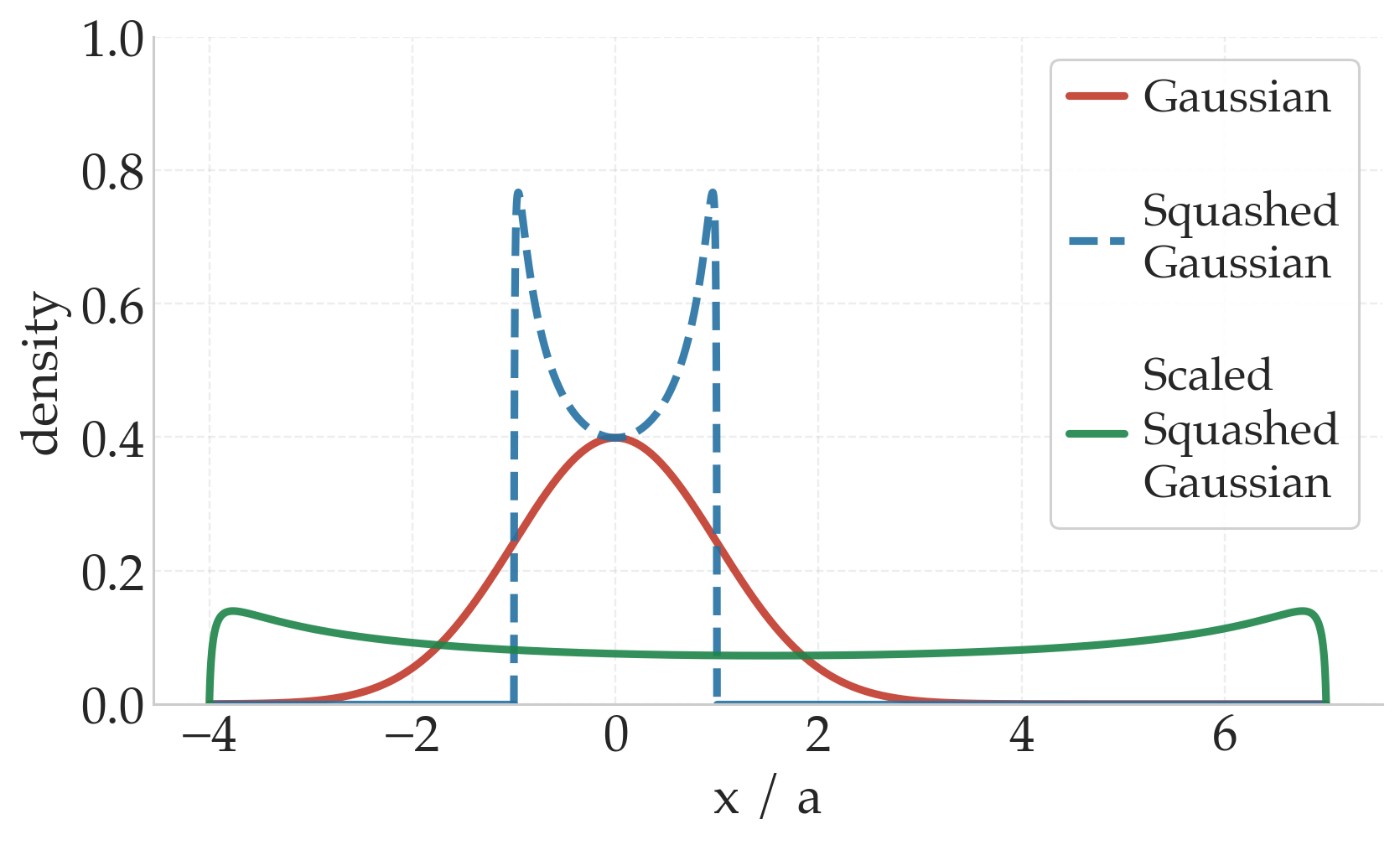}
(a) $\sigma_0 = 1.0$
\end{minipage}
\hfill
\begin{minipage}{0.45\textwidth}
\centering
\includegraphics[width=\linewidth]{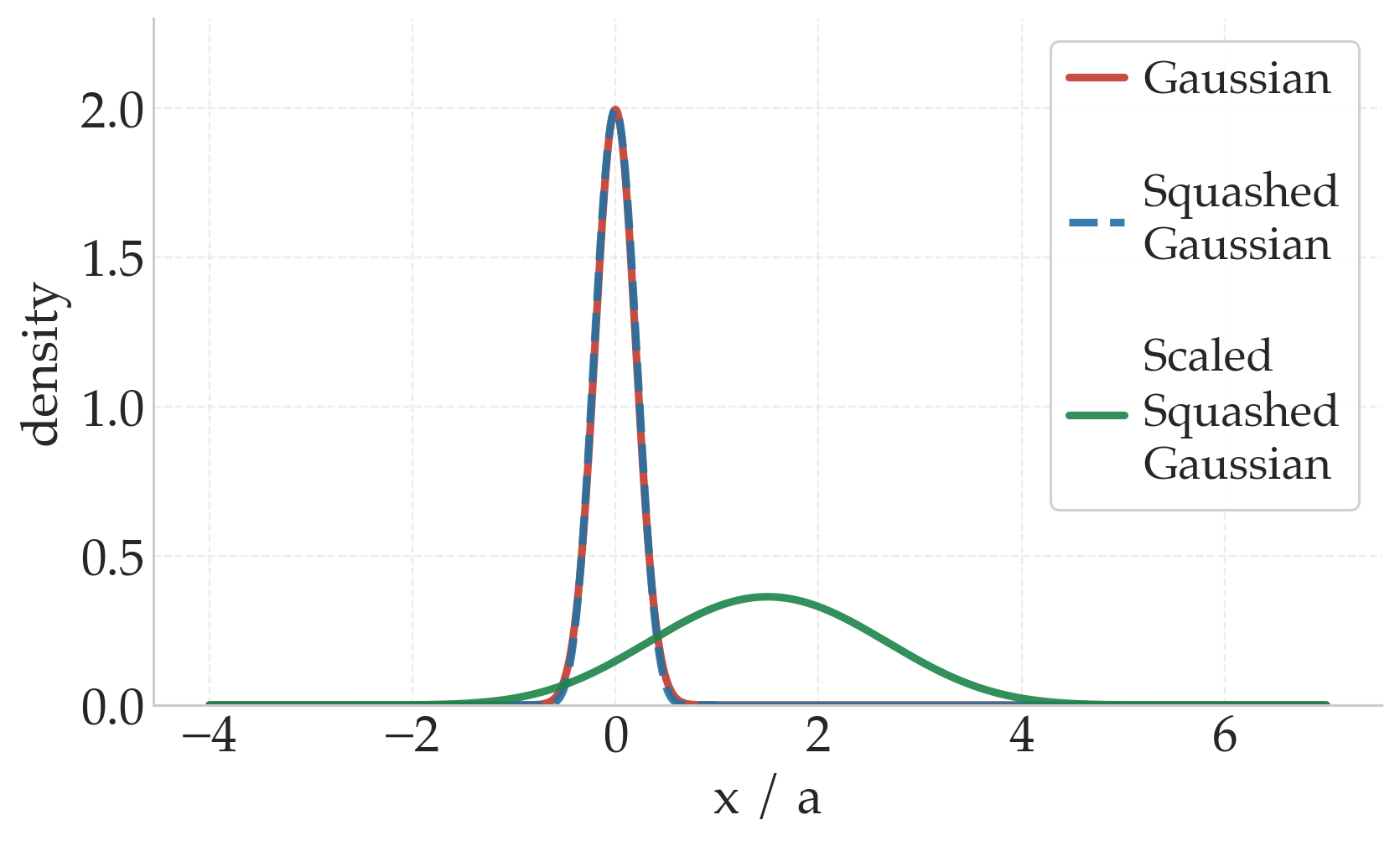}
(b) $\sigma_0 = 0.2$
\end{minipage}
\caption{Effect of initial standard deviation $\sigma_0$ on the squashed 
Gaussian exploration distribution. Large $\sigma_0$ leads to near-uniform 
exploration over the action range; small $\sigma_0$ concentrates 
exploration around the default joint configuration.}
\label{fig:noise_init}
\end{figure}

Even with correctly bounded actions, the shape of the initial exploration 
distribution is controlled by the initial policy standard deviation 
$\sigma_0$. For large values (e.g.\ $\sigma_0 = 1$), the squashed Gaussian 
places significant mass near the action bounds, resulting in near-uniform 
exploration over the admissible range. For small values (e.g.\ $\sigma_0 
= 0.15$), exploration concentrates around zero (i.e. around the default 
joint configuration), which we find leads to faster and more stable 
convergence. This behavior is illustrated in Fig.~\ref{fig:noise_init}.

To achieve this, the actor network ends with a linear layer mapping the 
last hidden feature $h \in \mathbb{R}^{d_h}$ to a $2d$-dimensional output 
$z = Wh + b$, which is then split into a mean head $z_\mu$ and a 
log-standard-deviation head $z_{\log\sigma}$, each of dimension $d$.

The mean head is initialized with small random weights and zero bias:
\begin{equation}
[W_\mu]_{ij} \sim \mathcal{N}(0, \epsilon),
\qquad
b_\mu = \mathbf{0}_d.
\end{equation}
This ensures that initial actions are close to zero, corresponding to 
the default joint configuration, while remaining weakly state-dependent. 
The small but nonzero weights also preserve gradient flow through the mean 
head from the very first update.

The log-standard-deviation head is initialized with a constant bias:
\begin{equation}
W_{\log \sigma} = \mathbf{0}_{d \times d_h},
\qquad
b_{\log \sigma} = \log(\sigma_0)\mathbf{1}_d.
\end{equation}
Setting the weights to zero makes the initial exploration scale 
state-independent and uniformly equal to $\sigma_0$ across all action 
dimensions, decoupling the exploration level from the network's hidden 
representation at initialization. Together, these two choices produce a 
policy centered at the default joint configuration with a controllable 
and well-calibrated exploration scale.

\subsection{Handling Timeout Transitions}
\label{sec:timeout}

SAC is derived for a discounted infinite-horizon objective, but simulation 
episodes are terminated artificially after a fixed number of steps for 
computational reasons. If these \emph{timeouts} are treated identically 
to true task failures, critic targets are biased: future value is 
incorrectly dropped at episode boundaries that do not correspond to 
genuine terminations. Correcting for this requires bootstrapping from the 
state immediately before the environment reset, rather than the post-reset 
state that the environment exposes by default.

In on-policy methods such as PPO, this issue can be handled approximately 
by evaluating the current value function at the timeout state. Because 
on-policy transitions are consumed immediately within the same iteration 
and then discarded, the value estimate used for bootstrapping is always 
consistent with the current policy, making this approximation acceptable 
in practice. SAC, by contrast, stores transitions in a replay buffer where 
they may be sampled repeatedly across many future updates. This introduces 
two compounding sources of staleness: first, the value function used for 
bootstrapping evolves continuously during training, so a bootstrap term 
precomputed at collection time becomes increasingly inaccurate as the 
critic is updated; second, and more fundamentally, SAC must evaluate the 
Q-function on actions sampled from the \emph{current} policy, not the 
policy that was active when the transition was collected. As training 
progresses and the policy changes, precomputed action samples become 
off-policy in a way that is inconsistent with the SAC objective, 
introducing bias that can destabilize training.

To address this, we modify the RSL-RL \texttt{VecEnv} wrapper, as well as the manager-based RL environment, to expose 
both the pre-reset observation $s_{t+1}^{\text{p}}$ and the post-reset 
observation $s_{t+1}$ through the \texttt{extras} interface. Let 
$b_t \in \{0,1\}^{N_e}$ and $d_t \in \{0,1\}^{N_e}$ denote the 
per-environment timeout and termination masks at time $t$, where $d_t$ 
encompasses both failures and timeouts. The corrected next observation is:
\begin{equation}
\tilde{s}_{t+1}
=
b_t \odot s_{t+1}^{\text{p}}
+
(1-b_t)\odot s_{t+1},
\end{equation}
and the replay buffer stores the tuple $(s_t, a_t, r_t, \tilde{s}_{t+1}, 
d_t, b_t)$. By storing $\tilde{s}_{t+1}$ rather than a precomputed 
bootstrap term, the value estimate and the action used for bootstrapping 
are both recomputed at training time using the current critic and policy, 
ensuring consistency throughout training. The bootstrap mask is 
$m_t = b_t + 1 - d_t$, giving the corrected SAC target:
\begin{equation}
\begin{aligned}
Q_{\bar{\theta}}^{\text{target}}
&=
r_t
+
\gamma m_t \mathbb{E}_{\tilde{s}_{t+1}\sim p}\left[
V_{\bar{\theta}}(\tilde{s}_{t+1})\right].
\end{aligned}
\end{equation}
All other optimization objectives are modified accordingly.

\subsection{$n$-Step Returns}
\label{sec:nstep}

One-step TD targets are noisy in rough-terrain tasks, where rewards are 
sparse over short horizons and a single misstep can cause termination. 
Following \citet{seo2025learningsimtorealhumanoidlocomotion} and 
\citet{seo2025fasttd3}, we augment the replay buffer to construct 
$n$-step returns on-the-fly, which accelerates reward propagation and 
improves training stability. Let $S_k := \prod_{j=0}^{k-1}(1 - d_{t+j})$, 
with $S_0 = 1$, denote the survival indicator over the $n$-step window. 
The masked $n$-step critic target is:
\begin{equation}
Q_{\bar{\theta}}^{\text{target}}
=
\sum_{k=0}^{n-1} \gamma^k S_k r_{t+k}
+
\gamma^n S_n V(\tilde{s}_{t+n})
+
\sum_{k=0}^{n-1} \gamma^{k+1} S_k b_{t+k}
V(\tilde{s}_{t+k+1}),
\end{equation}
where the expectation over the next state is omitted for brevity. The 
first term accumulates discounted rewards over $n$ steps, masking out 
transitions that occur after any termination. The second term bootstraps 
from the value function at step $t+n$, provided no termination occurred 
within the window. The third term handles timeout transitions consistently 
with Sec.~\ref{sec:timeout}: episode failures stop the return entirely, 
while timeouts allow bootstrapping from the corresponding pre-reset state.

Algorithm~\ref{alg:sac} summarizes the full training loop.

\begin{algorithm}[t]
    \caption{RSL-RL-SAC}
    \label{alg:sac}
    \begin{algorithmic}[1]
        \STATE \textbf{Initialize} actor $\pi_\phi$, critics $Q_{\theta_i}$ and targets $Q_{\bar\theta_i}$ ($i\in\{1,2\}$), buffer $\gD$, temperature $\alpha$
        \STATE Compute per-joint action bounds from soft joint limits (Sec.~\ref{sec:policy_param})
        \FOR{each learning iteration}
            \FOR{$N_s$ environment steps across $N_e$ parallel envs}
                \STATE Sample action $a_t \sim \pi_\phi(\cdot \mid s_t)$
                \STATE Step environments; observe $r_t$, $s_{t+1}$, done flag $d_t$, timeout flag $b_t$
                \STATE Build corrected next-state $\tilde{s}_{t+1}$ (Sec.~\ref{sec:timeout})
                \STATE Store $(s_t,\, a_t,\, r_t,\, \tilde{s}_{t+1},\, d_t,\, b_t)$ in $\gD$
            \ENDFOR
            \FOR{each gradient update step}
                \STATE Sample $n$-step mini-batch from $\gD$
                \STATE Compute $n$-step return target $y$ with timeout correction (Sec.~\ref{sec:nstep})
                \STATE \textbf{Critic update:} minimize $J_Q(\theta_i) = \mathbb{E}\bigl[(Q_{\theta_i}(s,a)-y)^2\bigr]$ for $i=1,2$
                \STATE \textbf{Temperature update:} minimize $J(\log\alpha) = \mathbb{E}\bigl[-\alpha\,(\log\pi_\phi(a\mid s)+\bar{\mathcal{H}})\bigr]$
                \STATE \textbf{Actor update} (every $p$ steps): maximize $J_\pi(\phi) = \mathbb{E}\bigl[\min_i Q_{\theta_i}(s,a') - \alpha\log\pi_\phi(a'\mid s)\bigr]$
                \STATE \textbf{Soft target update:} $\bar\theta_i \leftarrow \tau\theta_i + (1-\tau)\bar\theta_i$
            \ENDFOR
        \ENDFOR
    \end{algorithmic}
\end{algorithm}

\section{Experiments}
\label{sec:experiments}

\begin{figure}[h]
    \centering
    \begin{minipage}[t]{0.48\linewidth}
        \centering
        \includegraphics[width=\linewidth]{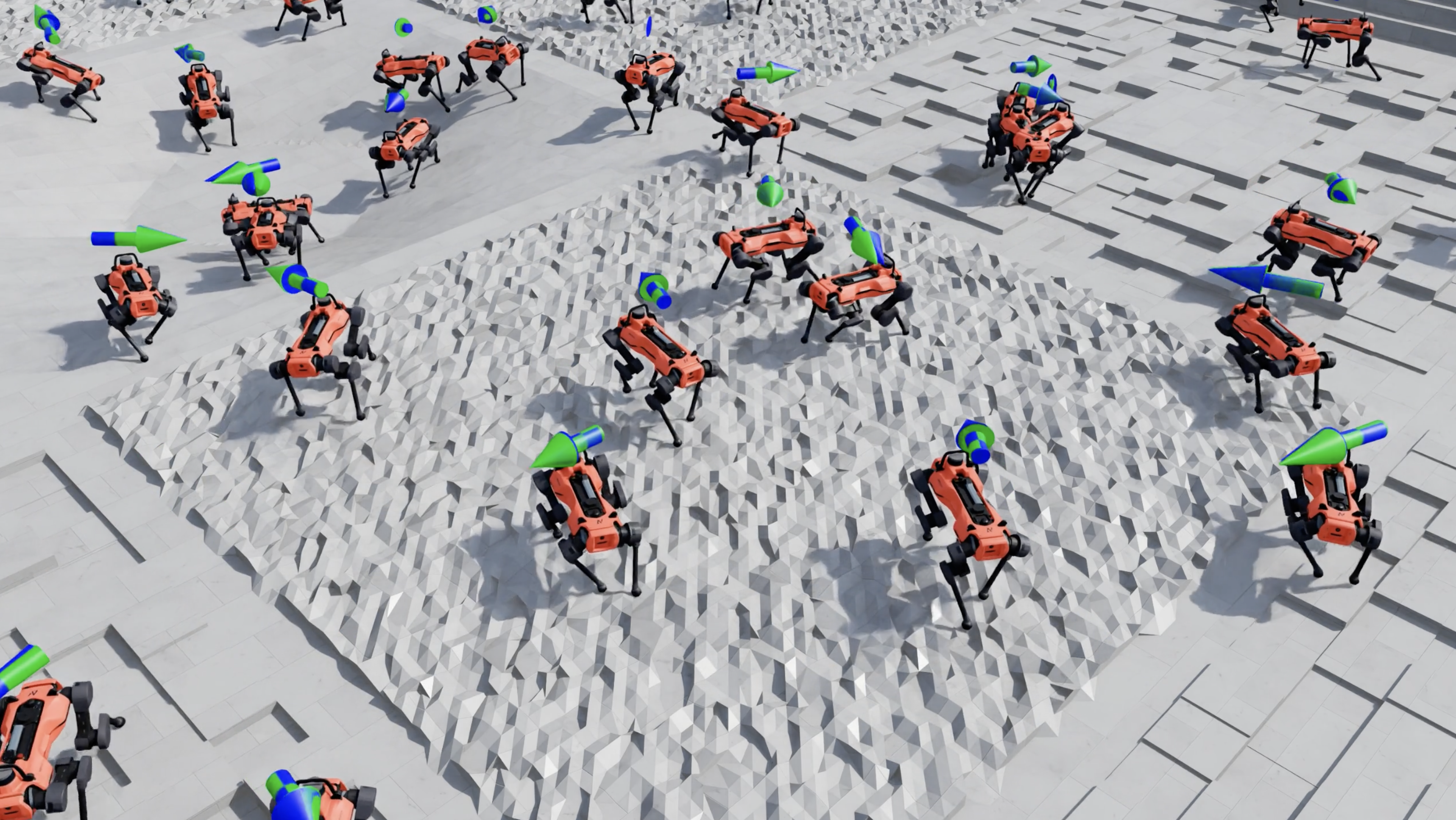}
    \end{minipage}
    \hfill
    \begin{minipage}[t]{0.48\linewidth}
        \centering
        \includegraphics[width=\linewidth]{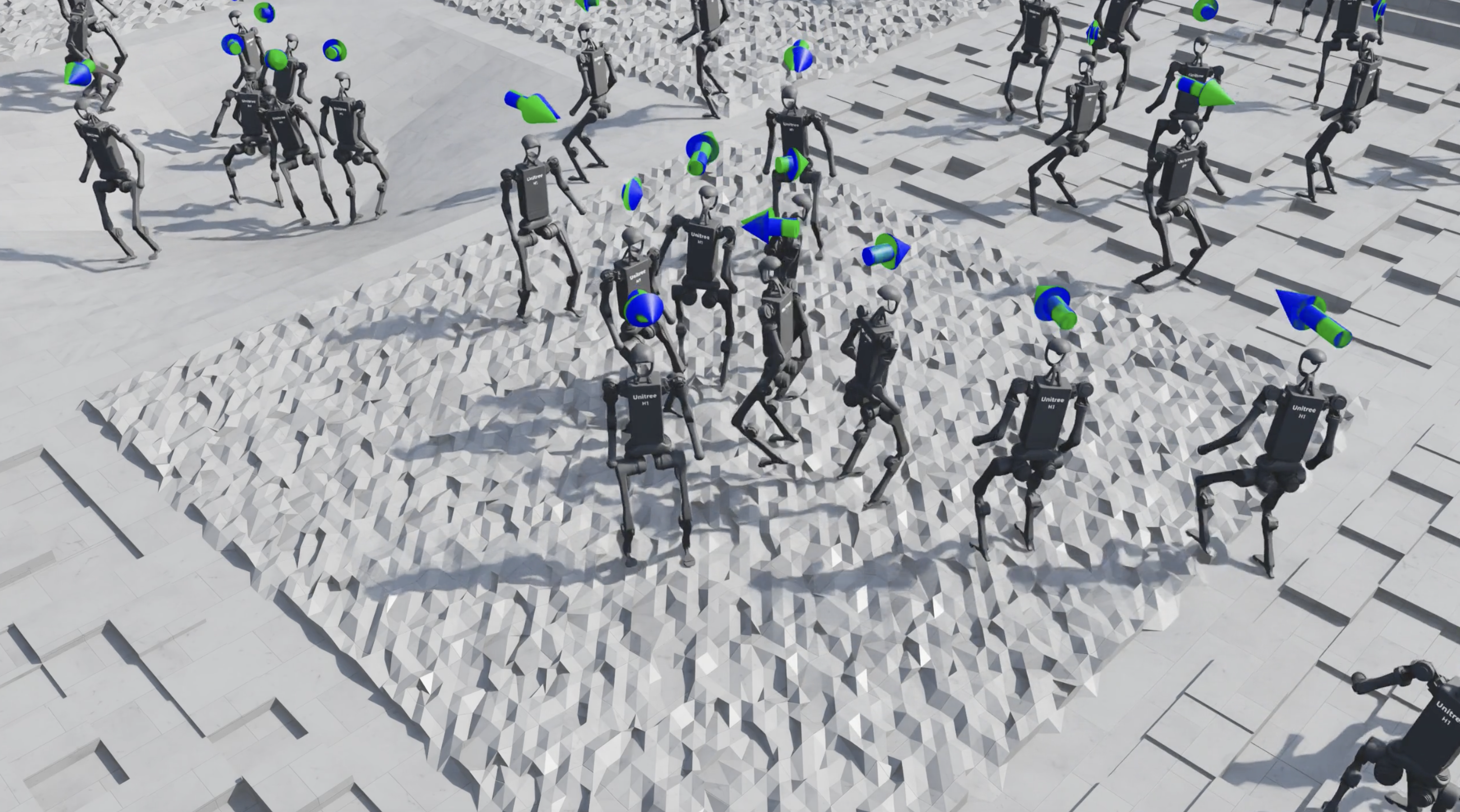}
    \end{minipage}
    \caption{Example of IsaacLab simulation environments used for evaluation: ANYmal D and Unitree H1 on rough terrain.}
    \label{fig:isaaclab_envs}
\end{figure}

\paragraph{Setup.}
We evaluate the proposed approach across a range of velocity tracking tasks 
on rough terrain, spanning both quadruped and humanoid platforms (Fig.~\ref{fig:isaaclab_envs}). To ensure 
a fair comparison with PPO, we adopt identical environment settings: $N_s = 
24$ rollout steps per environment and $N_e = 8192$ parallel environments. Crucially, we do not modify the 
reward functions and instead reuse those originally designed for PPO, which 
represents a more stringent evaluation setting than prior works that rely on 
task-specific reward engineering. Full hyperparameter details are provided 
in Sec.~\ref{supp:training_parameters}. Due to the high update-to-data (UTD) 
ratio required for stable SAC training~\citep{raffin2025isaacsim, 
seo2025learningsimtorealhumanoidlocomotion}, SAC performs significantly more 
gradient updates per unit of collected data than PPO, which contributes to 
the wall-clock time gap discussed below.

\begin{figure}[t]
\centering
\includegraphics[width=\linewidth]{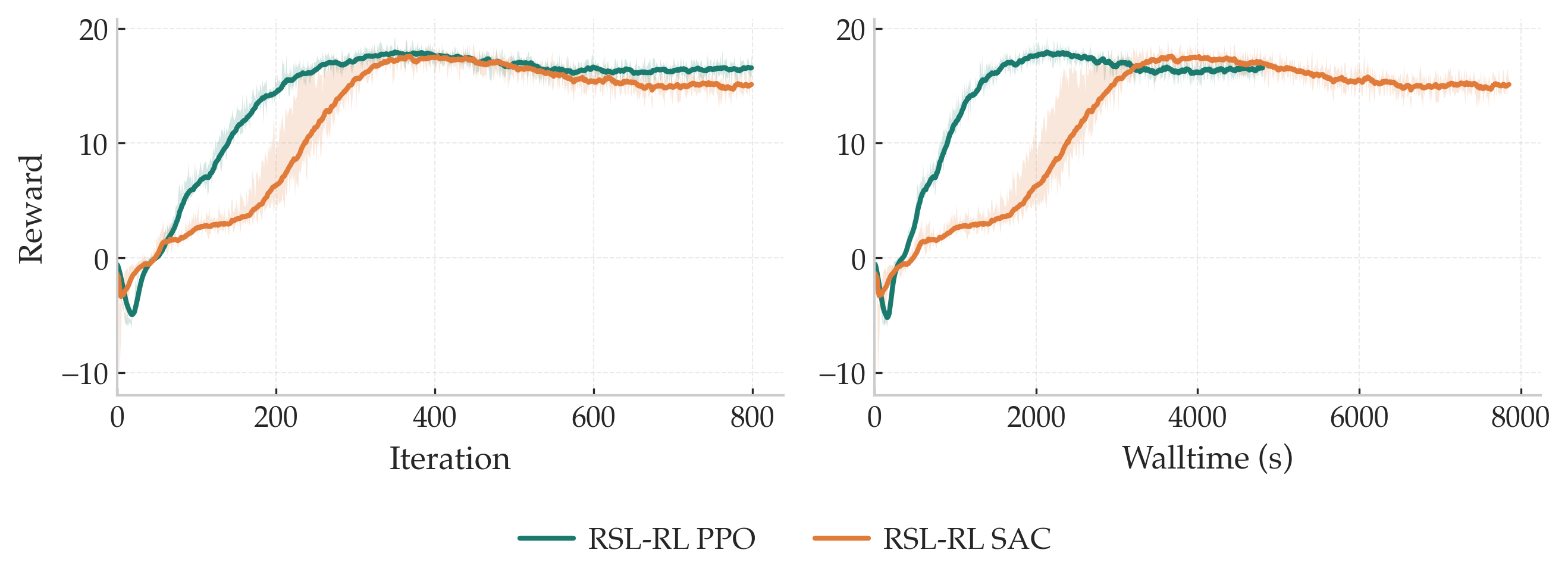}
\caption{Anymal D velocity tracking on rough terrain. Left: reward vs.\
training iterations. Right: reward vs.\ wall-clock time.}
\label{fig:anymal_d_rough}
\end{figure}

\begin{figure}[t]
\centering
\includegraphics[width=\linewidth]{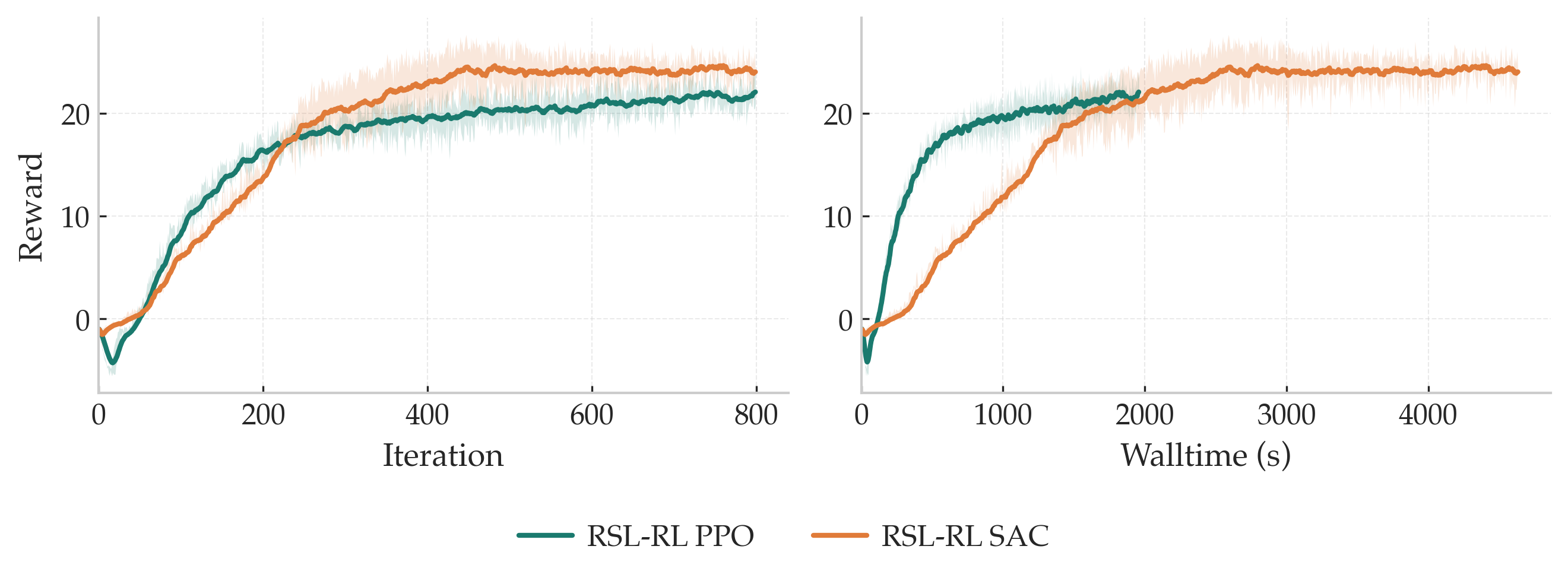}
\caption{Unitree A1 velocity tracking on rough terrain. Left: reward vs.\
training iterations. Right: reward vs.\ wall-clock time.}
\label{fig:a1_rough}
\end{figure}

\begin{figure}[t]
\centering
\includegraphics[width=\linewidth]{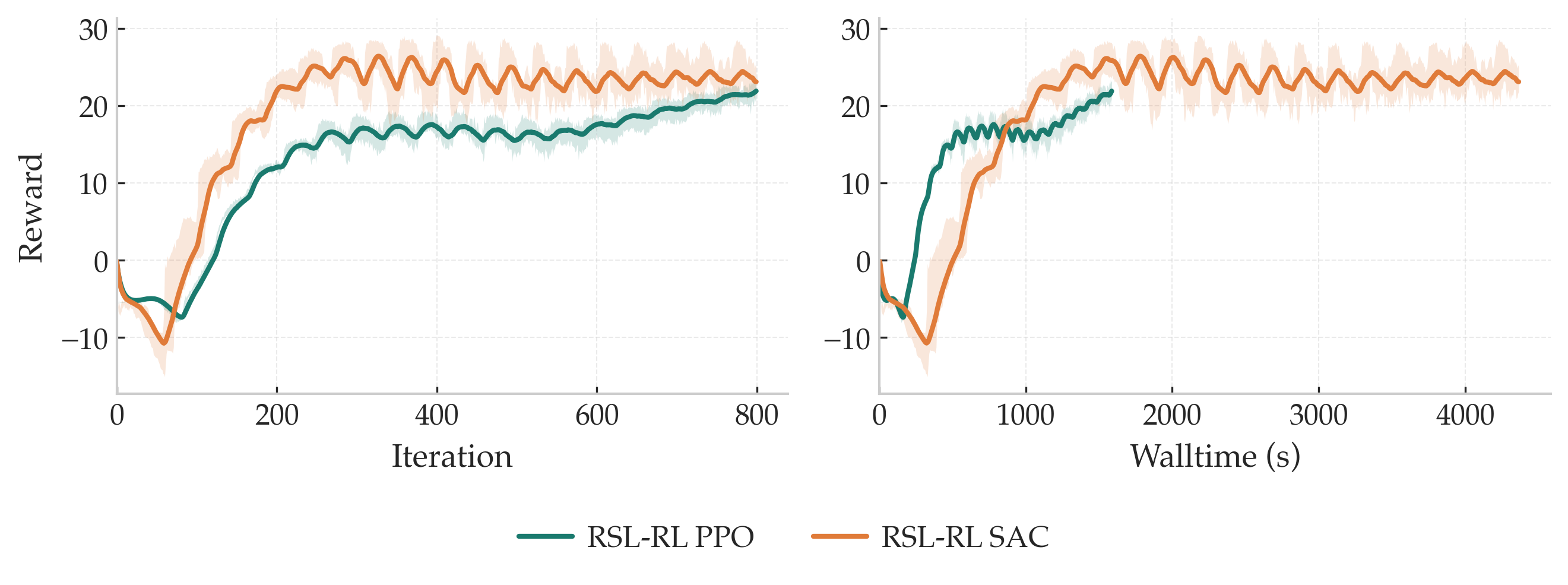}
\caption{Unitree H1 velocity tracking on rough terrain. Left: reward vs.\
training iterations. Right: reward vs.\ wall-clock time.}
\label{fig:h1_rough}
\end{figure}

\begin{figure}[t]
\centering
\includegraphics[width=\linewidth]{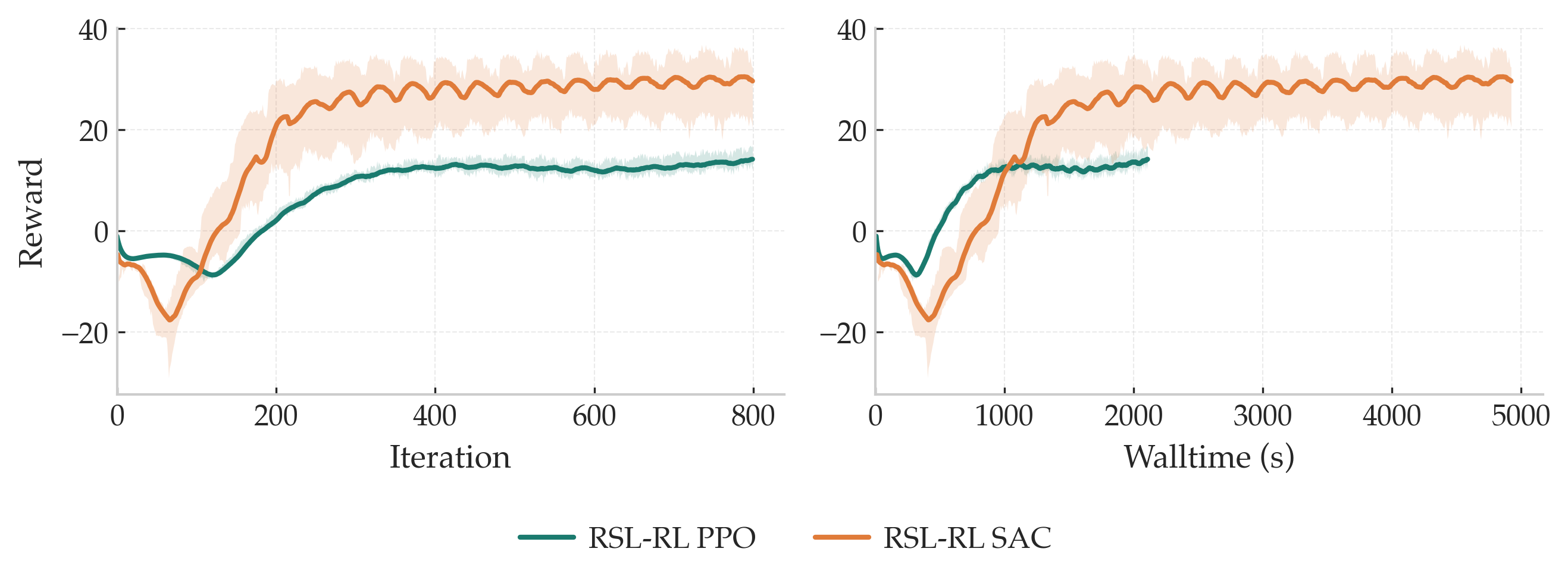}
\caption{Unitree G1 velocity tracking on rough terrain. Left: reward vs.\
training iterations. Right: reward vs.\ wall-clock time.}
\label{fig:g1_rough}
\end{figure}

As shown in Figs.~\ref{fig:anymal_d_rough}--\ref{fig:g1_rough}, the 
proposed modifications enable SAC to close the performance gap with PPO 
across all evaluated tasks (see also Sec.~\ref{supp:add_results} for 
additional platforms). On humanoid tasks, SAC surpasses PPO, which we 
attribute to the denser and more structured reward design used for these 
platforms (Sec.~\ref{supp:env_config}): in this setting, entropy 
maximization provides a meaningful exploration advantage that translates 
into higher final returns and faster convergence. Across all quadruped tasks, SAC converges to 
PPO-level performance with a single shared hyperparameter configuration, 
demonstrating that the proposed modifications are robust and transfer 
across platforms without per-task adjustment.

\section{Limitations}
\label{sec:limitations}

Closing the return gap does not resolve every challenge, and several
limitations remain relevant for practitioners considering SAC as a
replacement for PPO in massively parallel simulation.

\paragraph{Wall-clock time.}
Despite matching PPO in terms of cumulative reward, a significant
wall-clock time gap persists, and we argue it is structurally difficult
to eliminate. Several compounding factors contribute. First, SAC requires
larger network architectures to achieve comparable performance, increasing
the cost of every forward and backward pass. Second, each training
iteration involves multiple optimizers and loss terms, critic, actor,
and temperature, whereas PPO performs a single joint update, resulting
in lower per-iteration overhead. Third, off-policy replay exposes the
agent to a more diverse training distribution, which in turn requires a
higher update-to-data (UTD) ratio to extract a stable learning signal;
each environment step therefore triggers substantially more gradient
updates than in PPO. Finally, SAC lacks adaptive learning rate mechanisms
analogous to PPO's scheduling, requiring a learning rate roughly one
order of magnitude smaller to maintain stability
(Sec.~\ref{supp:training_parameters}), which further slows convergence
in wall-clock terms. Taken together, these factors mean that even when
SAC matches PPO in sample efficiency, it does so at greater computational
cost per unit time. Closing this gap would require either algorithmic
advances that reduce the UTD ratio needed for stability, or engineering
optimizations that
amortize the cost of the additional loss terms.

\paragraph{Other limitations.}
Entropy-driven exploration is not uniformly beneficial across tasks, and
understanding when it helps or hurts performance remains an open question.
We also observe higher variability across random seeds compared to PPO,
and in some tasks, notably G1 locomotion, high final rewards do not always
correspond to natural movement, suggesting that reward design may need to
be revisited for optimal behavior. Addressing these limitations will be
important for making SAC a robust drop-in replacement for PPO in the
massively parallel simulation setting.

\section{Conclusion}
\label{sec:conclusion}

This work demonstrates that the performance gap between SAC and PPO in 
massively parallel simulation can be closed through a principled set of 
modifications that address the root causes of SAC's instability in this 
setting. By correcting action space miscalibration, handling episode 
truncations consistently with the infinite-horizon training objective, 
and accelerating reward propagation through $n$-step returns, our approach 
enables SAC to achieve PPO-level performance across a broad range of legged 
locomotion tasks, from quadrupeds to humanoids, using a fixed set of  
hyperparameters and unmodified PPO reward functions. To 
facilitate further development and adoption, our implementation extends 
the baseline algorithm with Random Network Distillation 
(RND)~\citep{burda2018explorationrandomnetworkdistillation}, symmetry-based 
data augmentation~\citep{mittal2024symmetry}, and multi-GPU training support.

These results open a concrete path toward bridging simulation training and 
online reinforcement learning on real hardware. Because SAC operates 
off-policy, a policy pretrained in simulation can be fine-tuned directly 
on the physical robot using the same algorithm, without the massive 
environment parallelism that PPO requires. Investigating this 
sim-to-real fine-tuning workflow is a natural and promising direction for 
future work.

\subsubsection*{Acknowledgments}
\label{sec:acknowledgments}

This research was supported by the ETH AI Center and the Swiss National Science Foundation through the National Centre of Competence in Automation (NCCR automation).
In addition, we thank Clemens Schwarke for his technical support during the release.

\clearpage

\bibliography{references}
\bibliographystyle{iclr2026_conference}

\clearpage

\appendix
\section{Additional Results}
\label{supp:add_results}

\renewcommand{\thetable}{A\arabic{table}}
\renewcommand{\thefigure}{A\arabic{figure}}
\renewcommand{\theequation}{A\arabic{equation}}
In this section, we present additional results on velocity tracking tasks for Unitree Go1, Unitree Go2, and Anymal B. These results further support the findings reported in Sec.~\ref{sec:experiments}.

\begin{figure}[h]
\centering
\includegraphics[width=\linewidth]{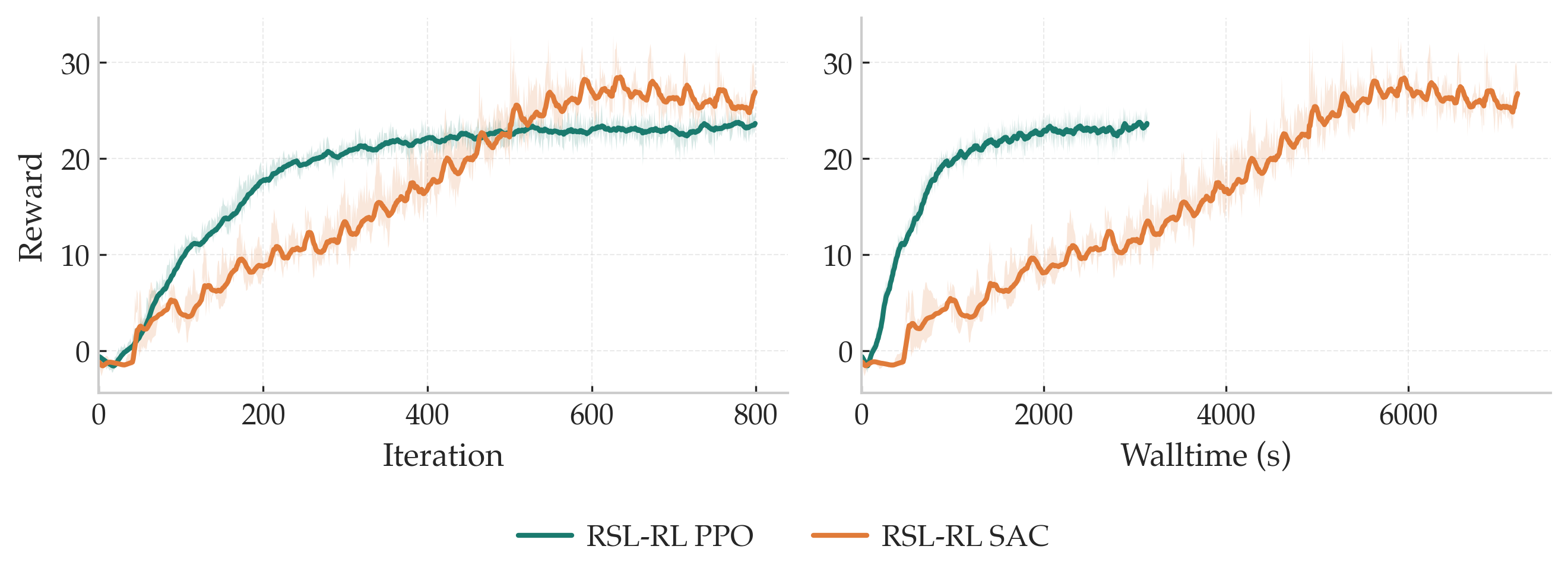}
\caption{Unitree Go1 Velocity Tracking on Rough Terrain}
\label{fig:go1_rough}
\end{figure}
\begin{figure}[h]
\centering
\includegraphics[width=\linewidth]{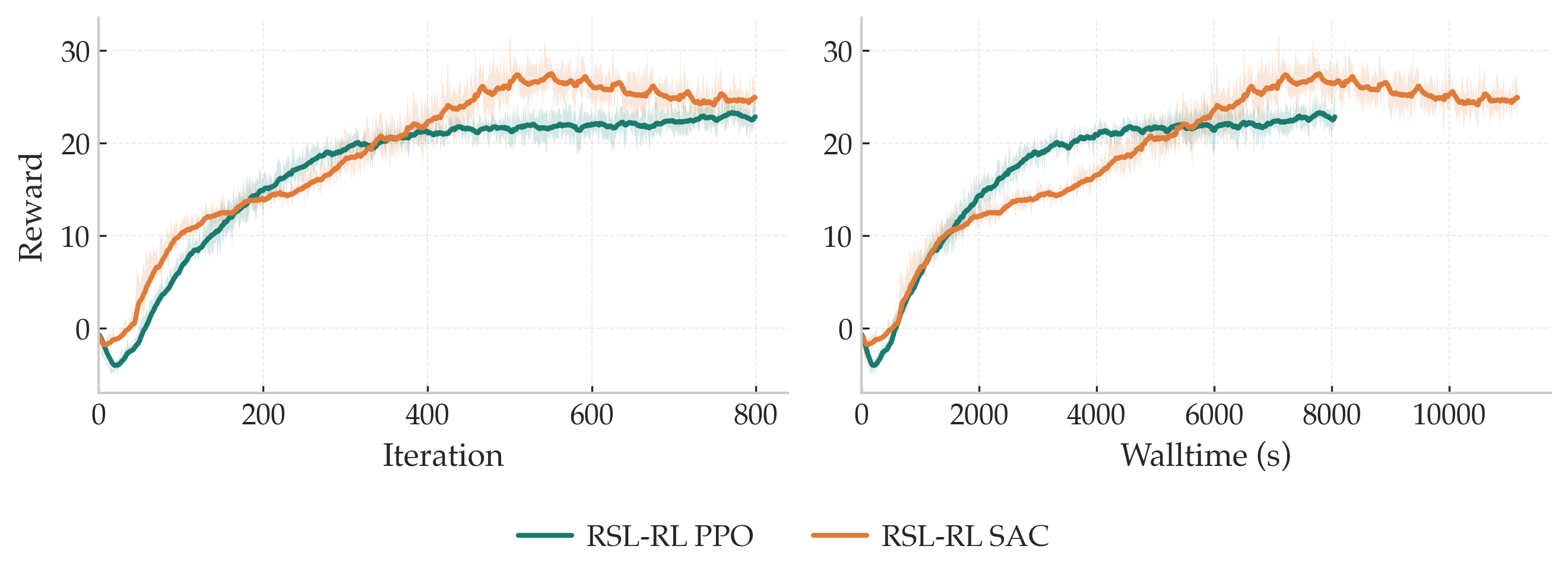}
\caption{Unitree Go2 Velocity Tracking on Rough Terrain}
\label{fig:go2_rough}
\end{figure}
\begin{figure}[h]
\centering
\includegraphics[width=\linewidth]{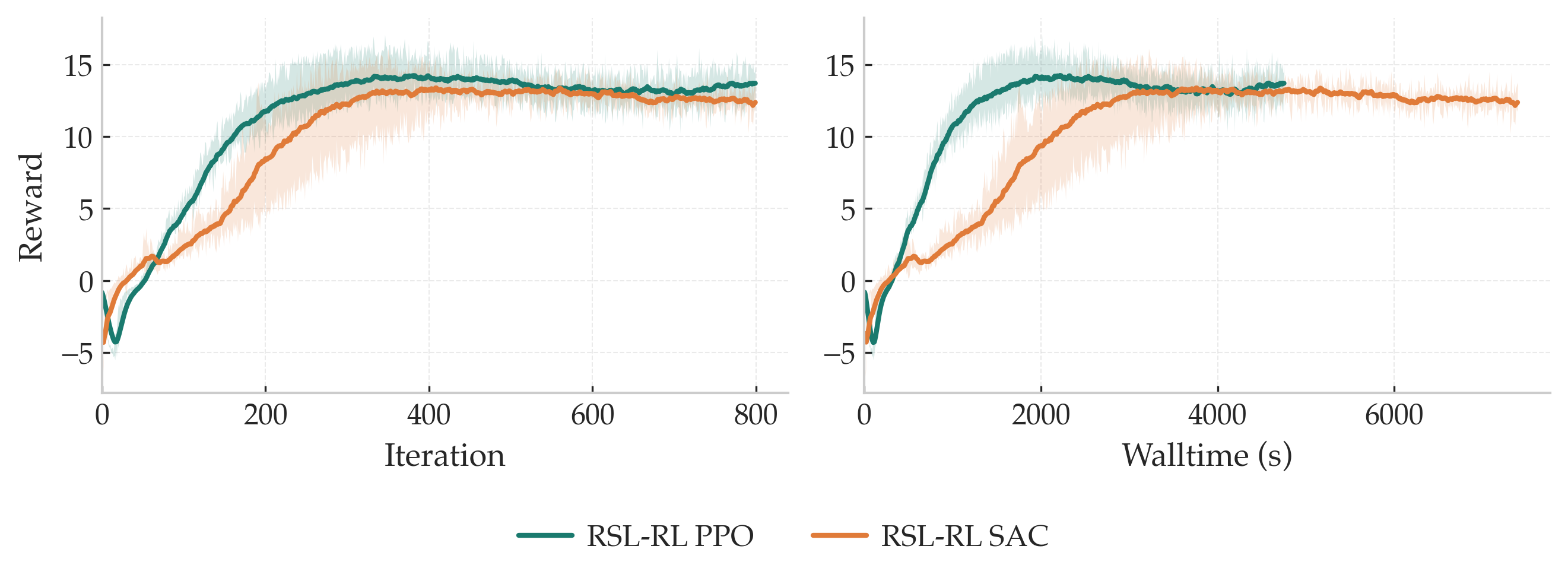}
\caption{Anymal B Velocity Tracking on Rough Terrain}
\label{fig:anymal_b_rough}
\end{figure}

\section{Environment Configuration}
\label{supp:env_config}
\renewcommand{\thetable}{B\arabic{table}}
\renewcommand{\thefigure}{B\arabic{figure}}
\renewcommand{\theequation}{B\arabic{equation}}
\subsection{Observation and action spaces}
\label{supp:obs_act_spaces}
The observation space for all the tasks mentioned in this paper can be found in the subsection in Tab. \ref{tab:obs_space_quadrupeds}--\ref{tab:obs_space_isaac_velocity_rough_g1_v0}. The action space is made of the joint position targets. The dimension of such space for each task can be seen in Tab. \ref{table:action_space}.

\begin{table}[]
    \centering
    \caption{Policy observation space for Anymal B/D, Unitree Go1/Go2/A1}
    \begin{tabular}{lcc}
    \toprule
    Entry & Symbol & Dimensions \\
    \midrule
    base linear velocity             & $v$        & 0:3 \\
    base angular velocity            & $\omega$   & 3:6 \\
    projected gravity                & $g$        & 6:9 \\
    commanded linear velocity (xy)   & $c_{xy}$   & 9:11 \\
    commanded angular velocity (z)   & $c_z$      & 11:12 \\
    joint positions                  & $q$        & 12:24 \\
    joint velocities                 & $\dot{q}$  & 24:36 \\
    previous actions                 & $a_{t-1}$  & 36:48 \\
    height scan                      & $h$        & 48:235 \\
    \bottomrule
    \end{tabular}
    \label{tab:obs_space_quadrupeds}
\end{table}

\begin{table}[]
    \centering
    \caption{Policy observation space for Unitree H1}
    \begin{tabular}{lcc}
    \toprule
    Entry & Symbol & Dimensions \\
    \midrule
    base linear velocity             & $v$        & 0:3 \\
    base angular velocity            & $\omega$   & 3:6 \\
    projected gravity                & $g$        & 6:9 \\
    commanded linear velocity (xy)   & $c_{xy}$   & 9:11 \\
    commanded angular velocity (z)   & $c_z$      & 11:12 \\
    joint positions                  & $q$        & 12:31 \\
    joint velocities                 & $\dot{q}$  & 31:50 \\
    previous actions                 & $a_{t-1}$  & 50:69 \\
    height scan                      & $h$        & 69:256 \\
    \bottomrule
    \end{tabular}
    \label{tab:obs_space_isaac_velocity_rough_h1_v0}
\end{table}

\begin{table}[]
    \centering
    \caption{Policy observation space for Unitree G1}
    \begin{tabular}{lcc}
    \toprule
    Entry & Symbol & Dimensions \\
    \midrule
    base linear velocity             & $v$        & 0:3 \\
    base angular velocity            & $\omega$   & 3:6 \\
    projected gravity                & $g$        & 6:9 \\
    commanded linear velocity (xy)   & $c_{xy}$   & 9:11 \\
    commanded angular velocity (z)   & $c_z$      & 11:12 \\
    joint positions                  & $q$        & 12:49 \\
    joint velocities                 & $\dot{q}$  & 49:86 \\
    previous actions                 & $a_{t-1}$  & 86:123 \\
    height scan                      & $h$        & 123:310 \\
    \bottomrule
    \end{tabular}
    \label{tab:obs_space_isaac_velocity_rough_g1_v0}
\end{table}

\begin{table}[]
    \centering
    \caption{Action space across robots}
    \begin{tabular}{lcc}
    \toprule
        Robot & Symbol & Dimensions \\
        \midrule
        Anymal B & $q^*$ & 12 \\
        Anymal D & $q^*$ & 12 \\
        Unitree A1 & $q^*$ & 12 \\
        Unitree Go1 & $q^*$ & 12 \\
        Unitree Go2 & $q^*$ & 12 \\
        Unitree G1 & $q^*$ & 37 \\
        Unitree H1 & $q^*$ & 19 \\
    \bottomrule
    \end{tabular}
    \label{table:action_space}
\end{table}

\subsection{Reward Functions}
\label{supp:reward_functions}

At each timestep, the reward is defined as a weighted sum of active components:
\begin{equation*}
    r = \sum_i w_i r_i .
\end{equation*}

The set of active terms and corresponding weights depends on the task and is reported in Tabs.~\ref{tab:reward_weights_anymal_bd}--\ref{tab:reward_weights_g1}. The following reward components are used across the considered tasks:

\begin{table}[]
    \centering
    \caption{Reward weights for Anymal B and Anymal D}
    \begin{tabular}{ll|ll}
    \toprule
    Symbol & Value & Symbol & Value \\
    \midrule
    $w_{v_{xy}}$ & $1.0$ & $w_{\omega_z}$ & $0.5$ \\
    $w_{v_z}$ & $-2.0$ & $w_{\omega_{xy}}$ & $-0.05$ \\
    $w_{q_\tau}$ & $-1.0\times10^{-5}$ & $w_{\ddot q}$ & $-2.5\times10^{-7}$ \\
    $w_{\dot a}$ & $-0.01$ & $w_{f_a}$ & $0.125$ \\
    $w_c$ & $-1.0$ & &  \\
    \bottomrule
    \end{tabular}
    \label{tab:reward_weights_anymal_bd}
\end{table}

\begin{table}[]
    \centering
    \caption{Reward weights for Unitree A1, Unitree Go1, Unitree Go2}
    \begin{tabular}{ll|ll}
    \toprule
    Symbol & Value & Symbol & Value \\
    \midrule
    $w_{v_{xy}}$ & $1.5$ & $w_{\omega_z}$ & $0.75$ \\
    $w_{v_z}$ & $-2.0$ & $w_{\omega_{xy}}$ & $-0.05$ \\
    $w_{q_\tau}$ & $-2.0\times10^{-4}$ & $w_{\ddot q}$ & $-2.5\times10^{-7}$ \\
    $w_{\dot a}$ & $-0.01$ & $w_{f_a}$ & $0.01$ \\
    \bottomrule
    \end{tabular}
    \label{tab:reward_weights_unitree_quads}
\end{table}

\begin{table}[]
    \centering
    \caption{Reward weights for Unitree H1}
    \begin{tabular}{ll|ll}
    \toprule
    Symbol & Value & Symbol & Value \\
    \midrule
    $w_{\mathrm{term}}$ & $-200.0$ & $w_{v_{xy}}$ & $1.0$ \\
    $w_{\omega_z}$ & $1.0$ & $w_{\omega_{xy}}$ & $-0.05$ \\
    $w_{\ddot q}$ & $-1.25\times10^{-7}$ & $w_{\dot a}$ & $-0.005$ \\
    $w_{f_a}$ & $0.25$ & $w_{f_s}$ & $-0.25$ \\
    $w_g$ & $-1.0$ & $w_{q^{\mathrm{ankle}}_{\mathrm{lim}}}$ & $-1.0$ \\
    $w_{q_d^{\mathrm{hip}}}$ & $-0.2$ & $w_{q_d^{\mathrm{arms}}}$ & $-0.2$ \\
    $w_{q_d^{\mathrm{torso}}}$ & $-0.1$ & & \\
    \bottomrule
    \end{tabular}
    \label{tab:reward_weights_h1}
\end{table}

\begin{table}[]
    \centering
    \caption{Reward weights for Unitree G1}
    \begin{tabular}{ll|ll}
    \toprule
    Symbol & Value & Symbol & Value \\
    \midrule
    $w_{\mathrm{term}}$ & $-200.0$ & $w_{v_{xy}}$ & $1.0$ \\
    $w_{\omega_z}$ & $2.0$ & $w_{\omega_{xy}}$ & $-0.05$ \\
    $w^{\mathrm{knee}}_{q_\tau}$ & $-1.5\times10^{-7}$ & $w^{\mathrm{ankle}}_{q_\tau}$ & $-1.5\times10^{-7}$ \\
    $w^{\mathrm{hip}}_{q_\tau}$ & $-1.5\times10^{-7}$ & $w_{\ddot q}$ & $-1.25\times10^{-7}$ \\
    $w_{\dot a}$ & $-0.005$ & $w_{f_a}$ & $0.25$ \\
    $w_{f_s}$ & $-0.1$ & $w_g$ & $-1.0$ \\
    $w^{\mathrm{ankle\text{-}pitch}}_{q_{\mathrm{lim}}}$ & $-1.0$ & $w^{\mathrm{ankle\text{-}roll}}_{q_{\mathrm{lim}}}$ & $-1.0$ \\
    $w_{q_d^{\mathrm{hip}}}$ & $-0.1$ & $w_{q_d^{\mathrm{arms}}}$ & $-0.1$ \\
    $w_{q_d^{\mathrm{fingers}}}$ & $-0.05$ & $w_{q_d^{\mathrm{torso}}}$ & $-0.1$ \\
    \bottomrule
    \end{tabular}
    \label{tab:reward_weights_g1}
\end{table}

\paragraph{Linear velocity tracking ($x,y$)}
\begin{equation*}
    r_{v_{xy}} = w_{v_{xy}} \exp\!\left(-\frac{\|c_{xy} - v_{xy}\|_2^2}{\sigma_{v_{xy}}^2}\right),
\end{equation*}
with $\sigma_{v_{xy}} = 0.5$.

\paragraph{Angular velocity tracking ($z$)}
\begin{equation*}
    r_{\omega_z} = w_{\omega_z} \exp\!\left(-\frac{\|c_z - \omega_z\|_2^2}{\sigma_{\omega_z}^2}\right),
\end{equation*}
with $\sigma_{\omega_z} = 0.5$.

\paragraph{Linear velocity penalty ($z$)}
\begin{equation*}
    r_{v_z} = w_{v_z}\|v_z\|_2^2.
\end{equation*}

\paragraph{Angular velocity penalty ($x,y$)}
\begin{equation*}
    r_{\omega_{xy}} = w_{\omega_{xy}}\|\omega_{xy}\|_2^2.
\end{equation*}

\paragraph{Joint torque penalty}
\begin{equation*}
    r_{q_\tau} = w_{q_\tau}\|\tau\|_2^2.
\end{equation*}

\paragraph{Joint acceleration penalty}
\begin{equation*}
    r_{\ddot q} = w_{\ddot q}\|\ddot q\|_2^2.
\end{equation*}

\paragraph{Action-rate penalty}
\begin{equation*}
    r_{\dot a} = w_{\dot a}\|a_t-a_{t-1}\|_2^2.
\end{equation*}

\paragraph{Feet air-time reward}
\begin{equation*}
    r_{f_a} = w_{f_a}\,t_{f_a},
\end{equation*}
where $t_{f_a}$ is the task-specific air-time statistic (quadruped vs. biped version in code).

\paragraph{Undesired contact penalty}
\begin{equation*}
    r_c = w_c\,c_u,
\end{equation*}
where $c_u$ counts contacts above threshold on undesired bodies.

\paragraph{Flat-orientation penalty}
\begin{equation*}
    r_g = w_g\,\|g_{xy}\|_2^2.
\end{equation*}

\paragraph{Joint-limit penalty}
\begin{equation*}
    r_{q_{\mathrm{lim}}}
    = w_{q_{\mathrm{lim}}}
    \sum_{j}
    \left(
        [\,q_j - q^{\max}_{j,\mathrm{soft}}\,]_+
        +
        [\,q^{\min}_{j,\mathrm{soft}} - q_j\,]_+
    \right),
\end{equation*}
where $[x]_+ = \max(x,0)$, and $q^{\min}_{j,\mathrm{soft}}, q^{\max}_{j,\mathrm{soft}}$ are the soft lower/upper limits for joint $j$.

\paragraph{Feet-slide penalty}
\begin{equation*}
    r_{f_s}
    =
    w_{f_s}
    \sum_{f\in\mathcal{F}}
    \left\|
        v^{w}_{f,xy}
    \right\|_2
    \,\mathbb{I}\!\left(
        \max_t \left\|F^{w}_{f}(t)\right\|_2 > 1.0
    \right),
\end{equation*}
where $\mathcal{F}$ denotes the set of feet considered in the task, $f$ is the foot index, and $w_{f_s}$ is the corresponding reward weight. Moreover, $v^{w}_{f,xy}\in\mathbb{R}^2$ denotes the linear velocity of foot $f$ in the world frame projected onto the horizontal plane, such that $\|v^{w}_{f,xy}\|_2$ is the planar sliding speed of that foot. The quantity $F^{w}_{f}(t)\in\mathbb{R}^3$ denotes the contact force acting on foot $f$ at time sample $t$ in the world frame, and $\max_t \|F^{w}_{f}(t)\|_2$ is the maximum contact-force magnitude over the sensor history used by the implementation. Finally, $\mathbb{I}(\cdot)$ is the indicator function.

\paragraph{Joint-deviation penalties}
\begin{equation*}
    r_{q_d} = w_{q_d}\|q-q_{0}\|_1,
\end{equation*}

\paragraph{Termination penalty}
\begin{equation*}
    r_{\mathrm{term}} = w_{\mathrm{term}} \, \mathbb{I}_{\mathrm{term}},
\end{equation*}
where $w_{\mathrm{term}}<0$ is a large negative constant and $\mathbb{I}_{\mathrm{term}}\in\{0,1\}$ is a binary indicator equal to $1$ if the environment terminates (failure) at the current step and $0$ otherwise.

\section{Training Parameters}
\label{supp:training_parameters}

\renewcommand{\thetable}{C\arabic{table}}
\renewcommand{\thefigure}{C\arabic{figure}}
\renewcommand{\theequation}{C\arabic{equation}}
The learning networks and algorithm are implemented in PyTorch 2.7.0 with CUDA 12.8 and trained on an NVIDIA RTX 3090 GPU. The hyperparameters used can be seen in Tab. \ref{table:sac_training}--\ref{table:ppo_training}. For the G1 and H1 platforms, task-specific hyperparameters are used: an entropy coefficient of 0.008 and 0.01 respectively for PPO, and a target entropy scale of 0.5 for SAC.

\begin{table}[h]
\centering
    \caption{SAC training parameters}
    \begin{tabular}{lcc}
    \toprule
        Parameter & Symbol & Value \\
        \midrule
        environments & $N_e$ & $8192$ \\
        step time seconds & $\Delta t$ & $0.02$ \\
        rollout steps per environment & $N_s$ & $24$ \\
        max iterations & $-$ & $800$ \\
        training start iteration & $-$ & $1$ \\
        replay buffer size & $|\mathcal{D}|$ & $5\times 10^6$ \\
        learning epochs & $-$ & $1$ \\
        mini-batches & $-$ & $200$ \\
        mini-batch size & $-$ & $8192$ \\
        actor learning rate & $\eta_{\pi}$ & $2\times 10^{-4}$ \\
        critic learning rate & $\eta_{Q}$ & $2\times 10^{-4}$ \\
        temperature learning rate & $\eta_{\alpha}$ & $2\times 10^{-5}$ \\
        discount factor & $\gamma$ & $0.97$ \\
        target smoothing coefficient & $\tau$ & $0.003$ \\
        initial temperature & $\alpha$ & $0.001$ \\
        automatic temperature tuning & $-$ & \texttt{True} \\
        target entropy scale & $-$ & $0.167$ \\
        max gradient norm & $-$ & $1.0$ \\
        policy update frequency & $-$ & $1$ \\
        multi-step return horizon & $n$ & $5$ \\
        actor observation normalization & $-$ & \texttt{True} \\
        critic observation normalization & $-$ & \texttt{True} \\
        actor hidden dimensions & $-$ & $[1024,\,512,\,256]$ \\
        critic hidden dimensions & $-$ & $[1024,\,512,\,256]$ \\
        activation & $-$ & \texttt{SiLU} \\
        layer normalization & $-$ & \texttt{False} \\
        initial action noise std & $\sigma_0$ & $0.15$ \\
    \bottomrule
    \end{tabular}
    \label{table:sac_training}
\end{table}

\begin{table}[h]
\centering
    \caption{PPO training parameters}
    \begin{tabular}{lcc}
    \toprule
        Parameter & Symbol & Value \\
        \midrule
        environments & $N_e$ & $8192$ \\
        step time seconds & $\Delta t$ & $0.02$ \\
        rollout steps per environment & $N_s$ & $24$ \\
        max iterations & $-$ & $800$ \\
        learning rate & $\eta$ & $1\times10^{-3}$ \\
        learning rate schedule & $-$ & adaptive \\
        learning epochs & $-$ & $5$ \\
        mini-batches & $-$ & $4$ \\
        discount factor & $\gamma$ & $0.99$ \\
        GAE parameter & $\lambda$ & $0.95$ \\
        clip range & $\epsilon$ & $0.2$ \\
        entropy coefficient & $-$ & $0.005$ \\
        value loss coefficient & $-$ & $1.0$ \\
        clipped value loss & $-$ & \texttt{True} \\
        normalize advantage per mini-batch & $-$ & \texttt{False} \\
        KL divergence target & $-$ & $0.01$ \\
        max gradient norm & $-$ & $1.0$ \\
        actor observation normalization & $-$ & \texttt{False} \\
        critic observation normalization & $-$ & \texttt{False} \\
        actor hidden dimensions & $-$ & $[512,\,256,\,128]$ \\
        critic hidden dimensions & $-$ & $[512,\,256,\,128]$ \\
        activation & $-$ & \texttt{ELU} \\
        initial action noise std & $\sigma_0$ & $1.0$ \\
    \bottomrule
    \end{tabular}
    \label{table:ppo_training}
\end{table}

\end{document}